\documentclass[10pt,a4paper,twocolumn]{article}

\usepackage[left=1.6cm,right=1.6cm,top=1.6cm,bottom=1.9cm]{geometry}
\usepackage{newpxtext,newpxmath} % sleek serif text/math
\usepackage[scaled=0.92]{DejaVuSansMono}
\usepackage[T1]{fontenc}
\usepackage[utf8]{inputenc}
\usepackage{microtype}
\usepackage{xcolor}
\usepackage{hyperref}
\usepackage{xcolor}
\definecolor{darkblue}{HTML}{0B3B8C}
\hypersetup{colorlinks=true,linkcolor=black,citecolor=darkblue,urlcolor=darkblue}
\usepackage{graphicx}
\usepackage{subcaption}
\usepackage{booktabs}
\usepackage{enumitem}
\usepackage{amsmath}
\usepackage{fontawesome5} % 
\usepackage{csquotes}
\usepackage{titlesec}
\usepackage{titling}
\usepackage{ragged2e}
\usepackage[labelfont=bf,font=small]{caption}
\usepackage[nameinlink,capitalize,noabbrev]{cleveref}
\crefformat{figure}{\textbf{#2Figure~#1#3}}

\titlespacing\section{0pt}{8pt plus 2pt minus 2pt}{6pt plus 2pt minus 2pt}
\titlespacing\subsection{0pt}{6pt plus 2pt minus 2pt}{4pt plus 1pt minus 1pt}
\titleformat{\section}{\large\bfseries}{\thesection}{0.6em}{}
\titleformat{\subsection}{\normalsize\bfseries}{\thesubsection}{0.6em}{}

\usepackage[normalem]{ulem}

\setlist[itemize]{itemsep=2pt, topsep=2pt, parsep=0pt, partopsep=0pt}

\pretitle{
  \begin{center}
  \rule{\linewidth}{0.75pt}
  \par\vspace{12pt}
  \fontsize{14}{16}\selectfont\bfseries
}
\posttitle{\par\vspace{4pt}\rule{\linewidth}{0.75pt}\end{center}}

\preauthor{%
  \par\vspace{0pt}%
  \begin{center}
    {\small \faGithub\ \href{https://github.com/joey-david/latent-correctness-probe}{\uline{\texttt{github.com/joeydavid/latent-correctness-probe}}}}
  \end{center}
  \vspace{0pt}%
  \begin{center}\normalsize
}
\postauthor{\par\vspace{0pt}\end{center}}

\title{Temporal Predictors of Outcome in Reasoning Language Models}
\author{
  \textbf{Joey David} \\[1pt]
  Independent Researcher \\[1pt]
  joeydhp@protonmail.com
}
\date{}

\begin{document}
\maketitle
\vspace{10mm}

% ---------- Abstract ----------
\renewenvironment{abstract}{
  \begin{center}
  \begin{minipage}{0.9\columnwidth}
  \centering{\large\bfseries\abstractname}
  \par\vspace{14pt}
  \justifying
  \ignorespaces
}{
  \end{minipage}
  \end{center}
}

\begin{abstract}
The chain-of-thought (CoT) paradigm uses the elicitation of step-by-step rationales as a proxy for reasoning, gradually refining the model’s latent representation of a solution. However, it remains unclear just how early a Large Language Model (LLM) internally commits to an eventual outcome. We probe this by training linear classifiers on hidden states after the first $t$ reasoning tokens, showing that eventual correctness is highly predictable after only a few tokens, even when longer outputs are needed to reach a definite answer. We show that, for harder questions, a drop in predictive accuracy highlights a selection artifact: hard items are disproportionately represented in long CoTs. Overall, our results imply that for reasoning models, internal self-assessment of success tends to emerge after only a few tokens, with implications for interpretability and for inference-time control.
\end{abstract}

\par\vspace{10pt}

% ---------- 1. Introduction ----------
\section{Introduction}
\label{sec:intro}

Late 2024 and early 2025 saw the first large-scale implementations of CoT finetuning and prompting, notably in OpenAI's \textit{o1} \cite{openai2024openaio1card} and Deepseek's \textit{R1} \cite{deepseekai2025deepseekr1} models. This led to LLMs achieving unheard-of performance in cognitively demanding tasks, especially in mathematical problem-solving and commonsense reasoning. The mechanism behind these improvements consists in pushing the model to generate a step-by-step path to a solution, openly formulating calculations or deductions, possibly reflecting on possible mistakes that could be made along the way before settling on a final answer. CoT generations are not mechanistically different from \textit{regular} token generation; they are simply specialized to problem-solving. Thus, deciding whether a model's latent representation has already converged onto a definite answer, or whether all of the CoT remains necessary to reach it, is an important but non-trivial problem. A model able to assess when it is honing in on the right solution may enable early error detection, better calibration, or dynamic halting of reasoning to save computation \cite{mao2025escot}.

% ---------- Fig 1 (answer length vs accuracy) ----------
\begin{figure}[t]
  \vspace{1.2cm}
  \centering
  \makebox[\columnwidth][c]{\includegraphics[width=1.05\columnwidth]{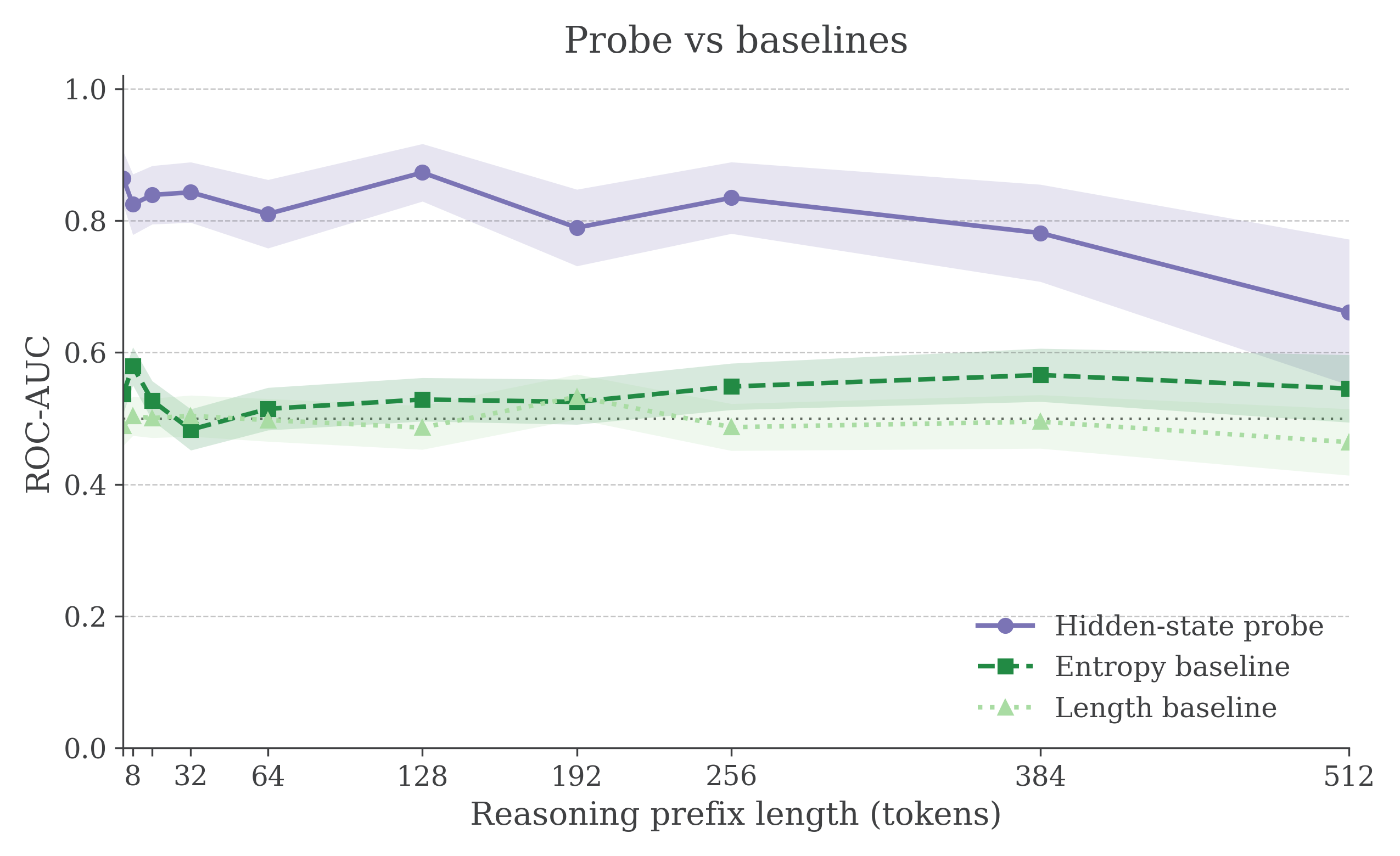}}
  \caption{Probe performance vs. entropy and length baselines, depending on CoT length.}
  \label{fig:len_acc}
\end{figure}

In this work, we try to assess whether a model is on a trajectory to give the correct solution \emph{throughout} a CoT rollout, before the final answer is produced. Prior work on “early correctness” probes has mostly targeted (i) question-only states or (ii) short-form QA, where surface correlations and entropy cues are strong. Here we consider the harder setting of math-focused CoT with variable-length reasoning, where such cues are weaker. Our experiments reveal a surprising result - a detectable correctness signal generally emerges very early into the reasoning process, even for longer CoT outputs. To ensure robust evaluation, we compare probe performance on the same set of problems across different prefix lengths $t$. Our analysis confirms that the model’s internal state truly encodes, to a significant extent, whether the answer will be correct well before outputting it.

Finally, we highlight how our approach differs from existing methods. Prior works on confidence estimation largely rely on post-hoc signals (the final answer logits or multiple sampled solutions) and added mechanisms at inference time. In contrast, we directly probe the model’s own hidden trajectory during reasoning, without requiring any extra decoding passes or fine-tuning of the model. Our results show that a simple linear probe is enough to tap into the model’s self-monitoring capabilities. By focusing on challenging long-form math problems, we uncover an early internal self-assessment that was previously elusive (earlier question-only probes struggled on math). In summary, our contributions include:

\begin{itemize}[leftmargin=1em,itemsep=2pt,topsep=2pt]
  \item \textbf{Probing during CoT generation.} We predict answer correctness directly from an LLM’s hidden state \emph{while} the chain-of-thought is being generated, not only after initial question or the final answer, using a fixed, auditable recipe.
  \item \textbf{Early, linearly decodable correctness signal.} We show that hidden states from very short prefixes (e.g. $t\!\approx\!4$) already contain linearly decodable information about eventual correctness, reaching AUCs up to \textbf{0.8+} on MATH-like data, which shows the model’s internal computation anticipates outcome likelihood before verbalization.
  \item \textbf{Difficulty-driven temporal effect.} We highlight that apparent performance “degradation” at later CoT steps is largely a \emph{temporal selection effect}: longer traces correspond to harder items, which both lengthens CoT and makes confidence signals sparser, rather than the hidden-state signal itself vanishing.
\end{itemize}

% ---------- 2. Related Work ----------
\section{Related Work}
\label{sec:related}

\subsection{Probing hidden activations for correctness and factuality.}

A growing line of work shows that LLM hidden states encode latent information about truthfulness and correctness that may not surface in its token-space output. \cite{cencerrado2025noanswer} extract the residual activation right after the question is processed by the model (i.e., before any answer token is produced) and train a linear probe to predict whether the model’s eventual answer will be correct. While able to reach above-chance success prediction on trivia-style QA, subsequent results of this work suggest that mathematical reasoning makes it harder to linearly decode internal confidence. Our work expands on this, and targets this gap by (i) focusing on math CoT and (ii) probing hidden states \emph{during} the reasoning trajectory, not only in the pre-answer state.

Similarly, \cite{gekhman2025insideout} introduce the \textit{Inside-Out} framework to quantify factual knowledge stored in hidden representations. They show that models frequently encode more correct facts internally than they actually express, with hidden representations yielding markedly higher recall than the generated text. Remarkably, they show that a model can internally embed a fact (with the correct answer being ranked highest in latent space) yet systematically fail to output it. These results strengthen the view that latent activations contain rich information signals, not fully exploited by standard decoding. We build on this observation in the context of CoT reasoning.

\subsection{Early stopping in CoT reasoning.}

Long CoT traces are expensive, and several methods try to stop decoding once the answer is effectively determined. ES-CoT \cite{mao2025escot} asks the model for its current answer at several points during the procedure; using the length of the resulting \textit{step-answer} compared to other step-answers as a criterion for convergence. While this approach yields substantial token savings with minimal accuracy loss, it relies entirely on output-space heuristics and does not exploit the model's internal signals. This behavior could lag behind the model's internal confidence, and the model might already carry a signal unique to the answer before it begins to repeat it in output. HALT-CoT \cite{laaouach2025haltcot} instead monitors the entropy of the next-answer distribution and stops when entropy falls below a threshold, cutting 15–30\% of tokens on GSM8K while keeping accuracy within about 0.4 points of full CoT. By contrast, our probe uses \emph{internal} signals: it aims to predict correctness from the hidden state \emph{before} the answer has converged. This potentially enables even earlier interventions - e.g., switching to self-reflection or to an alternative decoding strategy as soon as the internal probe flags low confidence; making our method complementary to ES-CoT and HALT-CoT. 

\subsection{Latent Reasoning and CoT}

Our work builds on chain-of-thought prompting \cite{wei2022chain}, which showed that providing structured intermediate steps allows LLMs to solve arithmetic and logical problems with much greater consistency than naïve prompting. Prior CoT research has noted that longer reasoning is not always better, models can “overthink” easy questions and introduce errors in late steps \cite{kojima2022large, zhou2023leastto}. This observation motivates internal progress monitoring: if the model’s activations early in the trajectory already suggest high eventual correctness, we could truncate or adapt the reasoning on the fly. To the best of our knowledge, our results are among the first to show that in complex multi-step math settings, partial CoT hidden states are greatly decodable for eventual correctness.

% ---------- 3. Method ----------
\section{Method}
\label{sec:method}

\subsection{Difficulty-balanced dataset}

% ---------- Fig 2 (balance) ----------
\begin{center}
\begin{minipage}{1.02\columnwidth}
  \centering
  \includegraphics[width=\linewidth]{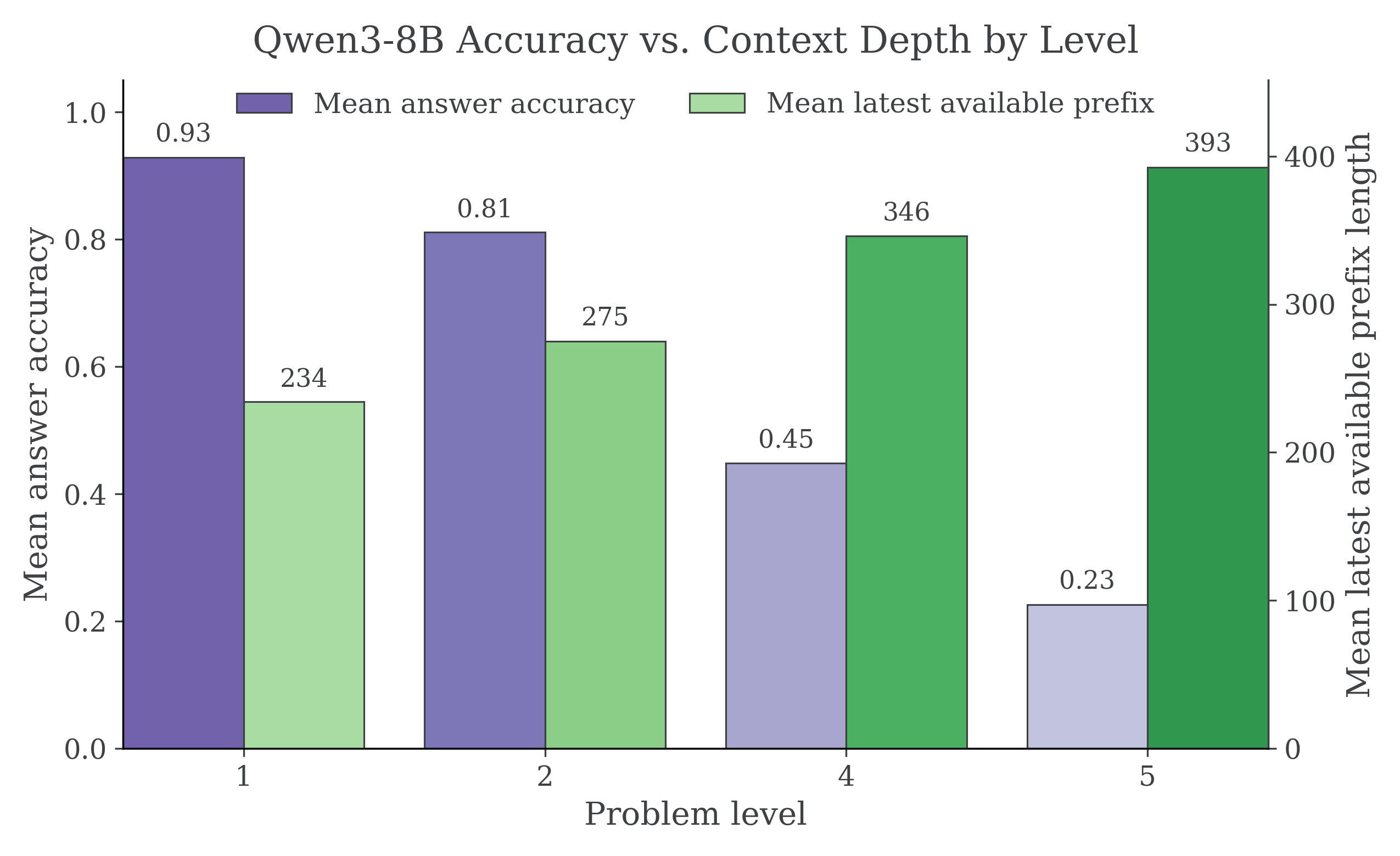}
  \captionof{figure}{Mean accuracy and CoT length per difficulty level.}
  \label{fig:raw-cf}
\end{minipage}
\end{center}

We have reasoning LLMs answer questions sampled from Hendrycks MATH \cite{hendrycksmath2021}. Thanks to its explicit labeling of difficulty by question, we separate questions into two main classes: levels 1 and 2 populate the \emph{easy} bucket, and levels 4 and 5 the \emph{hard} bucket. We draw 750 problems per bucket, yielding a total of 1{,}500 examples. This yields a well-rounded spread in reasoning length and complexity, with Qwen3-8B solving easy instances 85.1\% of the time, while hard ones drop to 32.1\%, giving a wide accuracy dynamic for probing latent confidence.

\subsection{Model Choice and Generation}

Because of limited access to computing resources, we restrict our testing to two 8B models:

\begin{itemize}[leftmargin=1em]
  \item \textbf{Qwen3-8B} - a model explicitly fine-tuned for reasoning. We enable its structured reasoning mode and initiate generation with the \verb|<think>| token.
  \item \textbf{Llama3.1-8B-Instruct} - an instruction-tuned model used to demonstrate that even models not tailored for reasoning exhibit a confidence signal above chance.
\end{itemize}

For both models, we choose greedy decoding and set \texttt{max\_new\_tokens}=512.

% ---------- Fig (PCA) ----------
\begin{center}
\begin{minipage}{1.02\columnwidth}
  \centering
  \includegraphics[width=\linewidth]{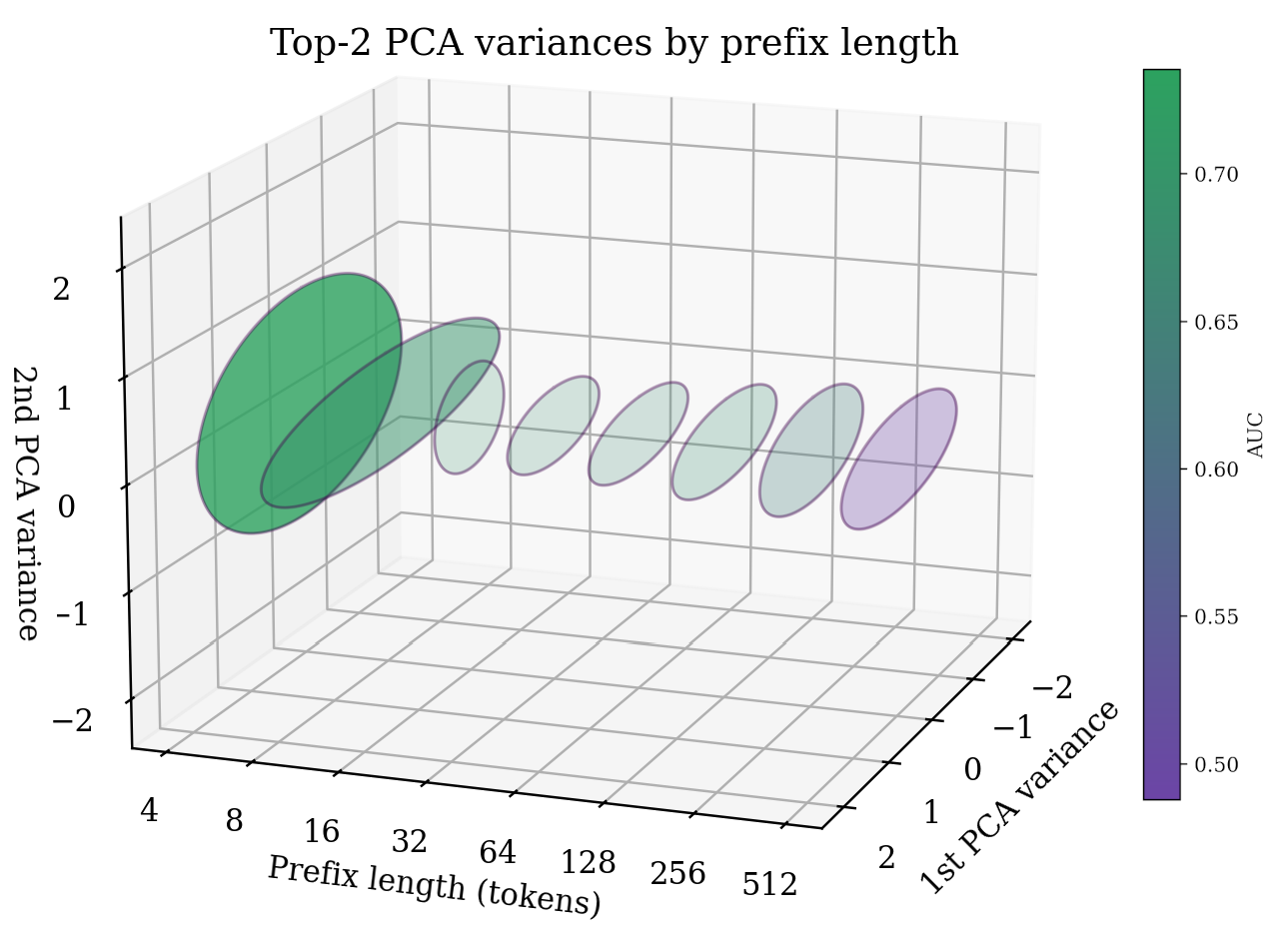}
  \captionof{figure}{Amplitude of the top-2 components of the PCA for Llama3.1-8B-Instruct.}
  \label{fig:pca}
\end{minipage}
\end{center}

\subsection{PCA and Linear Probe}
For every generated solution, we pool the final hidden states of the last four reasoning tokens at fixed prefix lengths $t \in \{4,8,16,32,64,128,192,256,384,512\}$. For each state, we then run PCA down to at most 128 components, then fit an $\ell_2$-regularized logistic regression probe to the resulting components. We stratify an 80/20 split and balance class weights to offset label skew. Evaluation reports accuracy and ROC-AUC alongside the class prior in the training fold.

% ---------- 4. Experiments ----------
\section{Experiments and Results}
\label{sec:experiments}

\subsection{Experimental setup}
All reported numbers come from the balanced MATH split described in \cref{sec:method}. Because long reasoning chains are rarer, the effective sample size drops from 1{,}500 prefixes at $t{=}4$ to 566 at $t{=}512$.

\subsection{Prefix-wise probe accuracy}

In a similar fashion to \cite{cencerrado2025noanswer}, we find that the probe is already highly predictive after four reasoning tokens: \cref{fig:diff_curves} shows an average ROC-AUC of \textbf{0.84} and an accuracy of \textbf{0.76} at $t{=}4$. The ROC-AUC and accuracy of the probe remain high - around \textbf{0.8} - well into the 100 token mark. Longer traces yield noisier statistics because fewer examples survive; beyond $t{=}256$, the label distribution becomes skewed toward incorrect answers as only genuinely hard questions remain.

\subsection{Difficulty and CoT-length effects}
Difficulty stratification confirms that the probe mirrors how quickly the base model commits to an answer. At $t{=}4$, the AUC is of 0.77 on easy items and 0.72 on hard items; by $t{=}32$ the gap widens (0.84 vs.\ 0.76). Hard problems therefore require longer prefixes before their internal state separates correct from incorrect trajectories. When we additionally condition on long chains (\textit{fixed $\leq\!256$}), the early AUC stays above 0.84, revealing that the information is already present even when we force every example to continue reasoning.

% ---------- Fig 1 (probe) ----------
\begin{center}
\begin{minipage}{1.02\columnwidth}
  \centering
  \includegraphics[width=\linewidth]{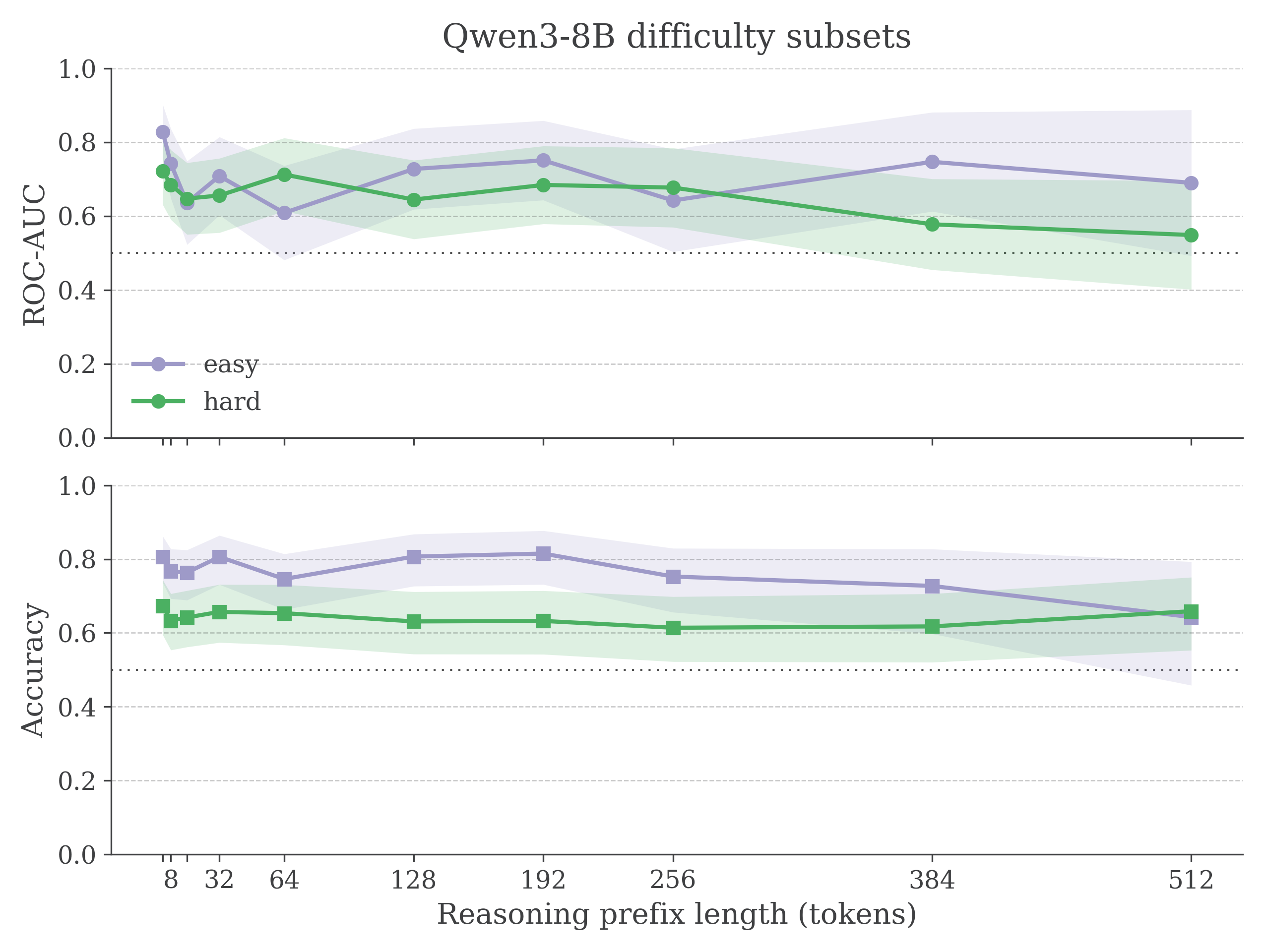}
  \captionof{figure}{ROC-AUC and Accuracy of the probe for easy and hard questions.}
  \label{fig:diff_curves}
\end{minipage}
\end{center}

The gradual decline in probe performance seen in \cref{fig:raw-cf} with increasing prefix length does not indicate that the model’s confidence deteriorates as reasoning progresses, but rather reflects a shift in the underlying sample composition. As $t$ grows, shorter and easier questions have already completed, leaving only the hardest examples in the remaining pool. Consequently, average probe accuracy drops because these residual items are intrinsically more difficult. This interpretation is confirmed by \cref{fig:diff_curves}, which shows that within each difficulty bucket, the probe’s accuracy remains nearly constant across prefixes (higher for easy problems and lower for hard ones), indicating that the hidden state consistently encodes both the model’s confidence and the inherent easiness of the question throughout generation.

\subsection{Comparison with output-space baselines}
Next-token entropy and prefix length provide weak signals relative to the hidden-state probe. Even when combined (\textit{entropy+length}) the best baseline AUC is 0.59 at $t{=}8$, more than 0.21 below the hidden-state probe at the same prefix. At the peak around $t{=}32$ the margin widens to 0.39. These gaps demonstrate that former uncertainty heuristics cannot explain the strong latent predictability we observe.

% ---------- 5. Discussion ----------
\section{Discussion and Limitations}
\label{sec:discussion}
Our results show that a linear probe needs only a few reasoning tokens before it predicts the eventual correctness of reasoning language models with high fidelity. The probe also offers a lightweight diagnostic for routing or self-monitoring systems that must decide when to re-run a solution, escalate to human review, or trigger reflective prompting.

The divergence between easy and hard items highlights a second takeaway: long chains mostly arise from inherently difficult questions rather than from meandering solutions. While we observe a modest drop in probe accuracy on the hardest slices, the carry-forward analysis shows that latent evidence for correctness is already present. The missing ingredient appears to be the model's ability to act on that knowledge: for difficult math problems, Qwen3-8B often recognizes its failing path but lacks a corrective policy. This gap aligns with recent evaluations of reflection-heavy reasoning models and points toward combining probes with targeted re-decoding policies.

Despite the strong empirical signal, several limitations are to consider. First, the study focuses on a single dataset and prompting template. This path of exploration would no doubt benefit from being ran on non-mathematical domains, which may exhibit weaker early commitment. Second, our leakage controls remove trivial answer exposures but cannot fully rule out subtle cues, such as partial numeric structure, that a linear classifier could exploit. Third, because of limited compute, long-prefix statistics are based on a few hundred examples, making the tail estimates sensitive to sampling noise and class imbalance. Finally, we only probe the final decoder layer; intermediate layers or non-linear probes might surface richer temporal dynamics.

Future work should replicate the pipeline across model sizes, modalities, and reasoning styles, and pair probe outputs with adaptive decoding strategies (e.g., selective continuation or tool-use triggers). Incorporating causal interventions, such as masking intermediate computations, resampling alternative continuations and pairing with non-greedy decoding could further strengthen the case of the probe capturing genuine self-assessment rather than simple stylistic markers.

% ---------- 6. Conclusion ----------
\section{Conclusion}

\label{sec:conclusion}
Linear probes applied to early chain-of-thought prefixes reveal that reasoning models internalize solution quality far earlier than their explicit answers. Through difficulty-balanced sampling, we show that this signal emerges within the first few reasoning tokens and remains stable across prefixes once problem difficulty is controlled. These results indicate that models internally track their own progress and that latent self-assessment precedes verbalized reasoning. Looking forward, such early correctness signals could enable adaptive inference strategies; like halting, rerouting, or self-reflection—based on the model’s own hidden confidence, and provide a new diagnostic path into understanding how reasoning unfolds inside large language models.

% ---------- References ----------


\begingroup
\small
\begin{thebibliography}{9}

\bibitem[Hendrycks et~al.]{hendrycksmath2021}
Dan Hendrycks, et~al. \\ 
\textit{Measuring Mathematical Problem Solving with the MATH Dataset.} 
In \textit{NeurIPS}, 2021.

\bibitem[OpenAI et~al.]{openai2024openaio1card}
OpenAI, et~al. \\ 
\textit{OpenAI o1 System Card.} 
arXiv:2412.16720, 2024.

\bibitem[DeepSeek-AI et~al.]{deepseekai2025deepseekr1}
DeepSeek-AI, et~al. \\ 
\textit{DeepSeek-R1: Incentivizing Reasoning Capability in LLMs via Reinforcement Learning.} 
arXiv:2501.12948, 2025.

\bibitem[Gekhman et~al.]{gekhman2025insideout}
Zorik Gekhman, et~al. \\ 
\textit{Inside-Out: Hidden Factual Knowledge in Language Models.} 
arXiv:2503.15299, 2025.

\bibitem[Mao et~al.]{mao2025escot}
Minjia Mao, et~al. \\ 
\textit{Early Stopping Chain-of-Thoughts in Large Language Models (ES-CoT).} 
arXiv:2509.14004, 2025.

\bibitem[Laaouach et~al.]{laaouach2025haltcot}
Yassine Laaouach, et~al. \\ 
\textit{HALT-CoT: Model-Agnostic Early Stopping for Chain-of-Thought Reasoning via Answer Entropy.} 
OpenReview manuscript, 2025.

\bibitem[Wei et~al.]{wei2022chain}
Jason Wei, et~al. \\ 
\textit{Chain-of-Thought Prompting Elicits Reasoning in Large Language Models.} 
In \textit{ICLR}, 2022. arXiv:2201.11903.

\bibitem[Cencerrado et~al.]{cencerrado2025noanswer}
Iván Vicente Moreno Cencerrado, et~al. \\ 
\textit{No Answer Needed: Predicting LLM Answer Accuracy from Question-Only Linear Probes.} 
arXiv:2509.10625 [cs.CL], 2025. https://doi.org/10.48550/arXiv.2509.10625

\bibitem[Kojima et~al.]{kojima2022large}
Takeshi Kojima, et~al. \\ 
\textit{Large Language Models are Zero-Shot Reasoners.} 
In \textit{NeurIPS}, 2022. arXiv:2205.11916.

\bibitem[Zhou et~al.]{zhou2023leastto}
Denny Zhou, et~al. \\ 
\textit{Least-to-Most Prompting Enables Complex Reasoning in Large Language Models.} 
arXiv:2205.10625, 2023.



\end{thebibliography}

\endgroup
\end{document}